# LUMIR: an LLM-Driven Unified Agent Framework for Multi-task Infrared Spectroscopy Reasoning

*Zujie Xie[a,†], Zixuan Chen[a,†], Jiheng Liang[a], Xiangyang Yu[a], Ziru Yu [b,*]*

[a] School of Physics, State Key Laboratory of Optoelectronic Materials and Technologies, Sun Yat-Sen University, Guangzhou, 510275, China

[b] School of electronics and communication engineering, Sun Yat-Sen University, Shenzhen, 518107, China

[†] Zujie Xie and Zixuan Chen contribute equally to the article.

* Correspondence should be addressed to Ziru Yu; Email: yuziru@mail.sysu.edu.cn

**Abstract:** Infrared spectroscopy enables rapid, non destructive analysis of chemical and material properties, yet high dimensional signals and overlapping bands hinder conventional chemometric methods. Large language models (LLMs), with strong generalization and reasoning capabilities, offer new opportunities for automated spectral interpretation, but their potential in this domain remains largely untapped. This study introduces LUMIR (LLM-driven Unified agent framework for Multi-task Infrared spectroscopy Reasoning), an agent based framework designed to achieve accurate infrared spectral analysis under low data conditions. LUMIR integrates a structured literature knowledge base, automated preprocessing, feature extraction, and predictive modeling into a unified pipeline. By mining peer reviewed spectroscopy studies, it identifies validated preprocessing and feature derivation strategies, transforms spectra into low dimensional representations, and applies few-shot prompts for classification, regression, and anomaly detection. The framework was validated on diverse datasets, including the publicly available Milk near-infrared dataset, Chinese medicinal herbs, Citri Reticulatae Pericarpium(CRP) with different storage durations, an industrial wastewater COD dataset, and two additional public benchmarks, Tecator and Corn. Across these tasks, LUMIR achieved performance comparable to or surpassing established machine learning and deep learning models, particularly in resource limited settings. This work demonstrates that combining structured literature guidance with few-shot learning enables robust, scalable, and automated spectral interpretation. LUMIR establishes a new paradigm for applying LLMs to infrared spectroscopy, offering high accuracy with minimal labeled data and broad applicability across scientific and industrial domains.

## 1. Introduction

Infrared spectroscopy has emerged as a versatile analytical modality across food quality assessment, pharmaceutical process monitoring, agricultural product evaluation, and environmental sensing. Its non destructive nature, rapid data acquisition, and minimal sample preparation have rendered IR spectroscopy particular attraction for real time, in situ analysis in both academic research and industrial practice[1]. Nonetheless, the intrinsically high dimensionality of IR spectra, coupled with complex and overlapping absorption bands, imposes formidable challenges for robust, generalizable interpretation. Traditional machine learning and deep learning approaches ranging from partial least squares regression and support vector machines to convolutional neural networks and Transformer based architectures have achieved high accuracy on curated datasets through extensive feature engineering, model calibration, and hyperparameter optimization[2]. However, these methods often require large, representative training sets to mitigate overfitting, and they frequently necessitate retraining or recalibration when confronted with novel sample matrices or different instrumentation. Such dependencies introduce substantial costs in data collection and annotation and impede rapid deployment in variable real world environments.

Recent advances in large language models (LLMs) have demonstrated powerful capabilities in few-shot learning, chain of thought reasoning, and cross domain generalization[3][4]. Some pioneering studies have begun to apply LLMs to the scientific field, and LLMs have gradually developed into LLM based scientific agents that can automatically perform key scientific research tasks from hypothesis generation, experimental design to data analysis and simulation[5][6]. Unlike general purpose LLMs, these specialized agents integrate domain knowledge, advanced tool sets, and powerful verification mechanisms, enabling them to handle complex data types, ensure reproducible results, and promote scientific breakthroughs. LLMs are becoming a general tool for connecting computational methods, experimental data, literature text resources, and domain expertise[7]. In recent work on spectral analysis and detection, large language models (LLMs) have been applied in two complementary ways. One approach treats an LLM as an interpreter: high dimensional spectral signals are preprocessed and serialized (for example into JSON), then presented to the model through task specific prompt engineering or in context examples so the model can extract features, highlight diagnostic bands, or produce quantitative estimates without updating model weights. A second approach finetunes an LLM (or LLM derived architecture) on formatted spectral sequences so the model directly maps spectra to analytical targets. For example, Zhu et al. (2024) finetuned a Babbage-002 model on subpixel decomposed NIR spectra encoded in JSON to predict the toxin content, reporting higher accuracy than 1-D and 2-D CNN baselines[8]. Liang et al. (2024) used a prompt driven LLM workflow to extract characteristic UV-NIR spectral bands for chemical oxygen demand prediction, showing that carefully designed prompts and in context reasoning can match or exceed traditional machine learning methods while substantially reducing the need for task specific parameter tuning[9].

In parallel, several related end-to-end frameworks have emerged in other scientific domains. For instance, Large Language Model-Informed X-ray Photoelectron Spectroscopy (XPS) analysis integrates GPT-based reasoning with curve-fitting procedures to assist in interpreting photoelectron spectra[10], but it focuses primarily on post-processing rather than building a generalized multi-task pipeline. SpecCLIP introduces a foundation model that aligns and translates stellar spectroscopic measurements across different astronomical instruments[11], emphasizing cross-instrument calibration instead of chemometric method selection. More recently, XASDAML delivers a machine-learning-based platform for X-ray absorption spectroscopy, covering dataset construction, modeling, and statistical analysis[12], though it relies on conventional ML workflows rather than LLM-driven agents. Compared with these works, our framework uniquely integrates literature-guided preprocessing, feature extraction, and multi-task reasoning into a single LLM-driven agent, enabling automated spectral analysis across diverse infrared datasets with enhanced adaptability and interpretability.

To address the limitations of traditional chemometrics, we propose a unified, LLM-driven agent framework that seamlessly integrates literature based method retrieval, automated spectral preprocessing, feature extraction, and multi-task reasoning within a single end-to-end pipeline. A structured knowledge base of peer reviewed IR spectroscopy publications is first queried via retrieval algorithms to identify relevant preprocessing and feature extraction methods for the specified research object. These method descriptions are then mapped to a predefined Python function librarycomprising Asymmetric Least Squares baseline correction, Savitzky-Golay smoothing, standard normal variate transformation, multiplicative scatter correction, normalization, detrending, principal component analysis (PCA)[13], partial least squares (PLS) regression[14], Nonnegative Matrix Factorization (NMF)[15], continuous wavelet transform[16], spectral derivatives, peak detection, and statistical summarization without exposing raw high dimensional spectral arrays to the LLM. The agent invokes the selected functions in sequence to transform the spectral data into structured feature sets, ensuring that each step remains scientifically grounded and reproducible. Figure 1 shows the basic structure of an LLM-Driven Unified Agent Framework for Multi-task Infrared Spectroscopy Reasoning.

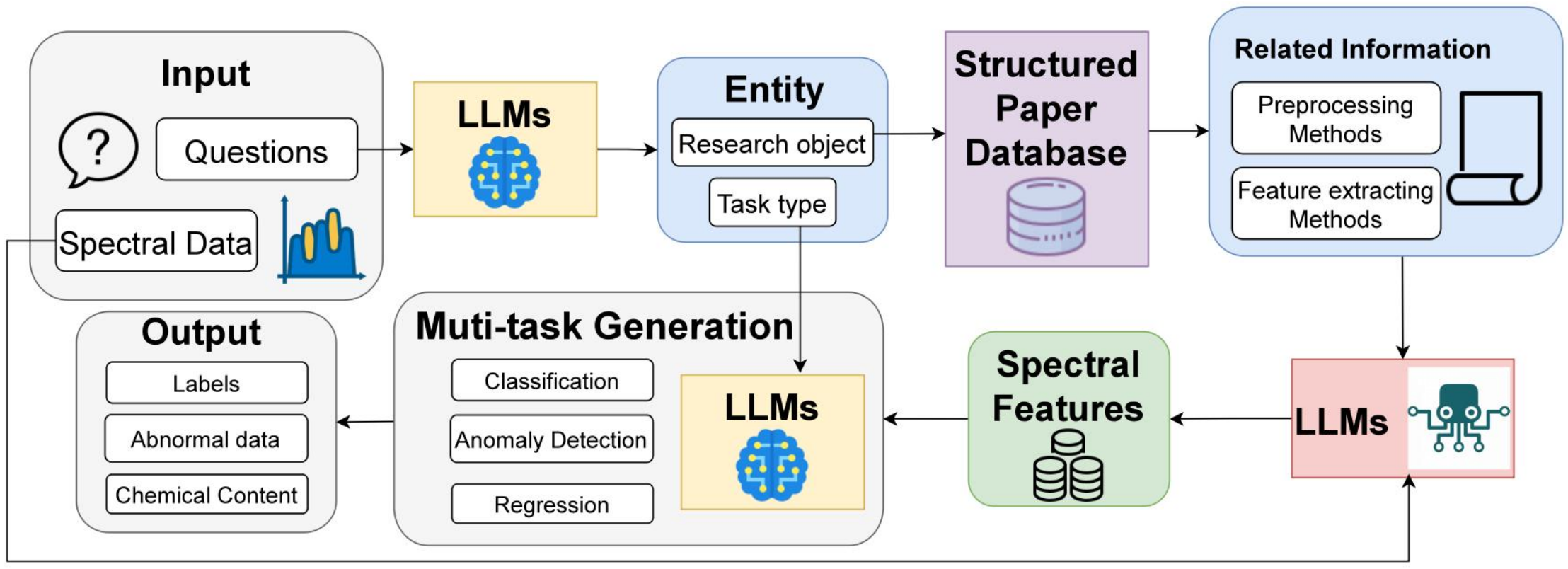

Figure 1. Basic structure of an LLM-Driven Unified Agent Framework for Multi-task Infrared Spectroscopy Reasoning

Building upon this automated preprocessing and feature extraction backbone, the framework integrates multi-task reasoning capabilities that allow the LLM to perform downstream analytical tasks such as classification, regression, and anomaly detection within a unified setting. For each task, the model is provided with few-shot examples drawn from the processed feature sets, and predictions are generated directly based on these contextual demonstrations. By embedding structured guidance and task specific prompting into the workflow, the proposed framework automates the selection and execution of chemometric methods and adapts flexibly to diverse analytical objectives, thereby offering an interpretable and scalable solution for infrared spectral analysis.

The key contributions of this work are:

1. End-to-end intelligent spectral analysis pipeline. We integrate a priori literature knowledge to select scientifically validated preprocessing and feature-extraction routines, then apply multi-turn LLM inference to execute diverse analytical tasks within a single workflow.

2. Unified multi-task framework. Our system concurrently performs classification, regression, and anomaly detection, overcoming the constraints of single-objective methods.

3. Few-shot prompting efficiency. By leveraging only minimal training data, the agent achieves high accuracy that consistently surpasses traditional machine learning models trained on equivalent sample sizes.

## 2. Materials and Method

### 2.1 Experimental materials

This study addressed three spectroscopic tasks, namely classification, anomaly detection, and regression, each of which was associated with specific sample sets and corresponding wavelength ranges. In the classification and anomaly detection experiments, three categories of materials were investigated. The classification task was designed to distinguish subclasses within a broader material category by assigning each spectrum to a specific subclass label, while the anomaly detection task was formulated as a binary problem to determine whether a sample belonged to a predefined subclass (True) or not (False).

The first dataset comprised milk samples with varying concentrations of lactose, measured within the 1100–2300 nm spectral range. This Milk NIR Spectral Dataset is an openly available benchmark provided by the NIRPyResearch initiative, intended to demonstrate the use of near-infrared spectroscopy for discriminating lactose content[17]. It includes 450 spectral scans obtained from nine distinct formulations, each representing a different mixture ratio of regular and lactose free milk. The second dataset contained three types of Chinese medicinal materials, namely Lonicerae Japonicae Flos, Lonicerae Flos, and their mixture, analyzed in the 900–1700 nm spectral range. The third dataset included Citri Reticulatae Pericarpium samples with storage durations ranging from one to eight years, with spectra also recorded between 900 and 1700 nm.

For the regression task, an internal wastewater dataset was used with the objective of predicting the chemical oxygen demand (COD) concentration of influent samples collected prior to treatment at a municipal wastewater treatment plant in Chongqing, China. In addition to this industrial dataset, two widely used public datasets, Tecator and Corn, were incorporated.

The Tecator dataset consists of absorbance spectra of chopped meat accompanied by reference measurements of fat, water, and protein, and the canonical regression problem focuses on predicting fat content[18]. The Corn dataset, made available by Eigenvector Research, contains eighty samples measured using three near-infrared spectrometers covering the 1100–2498 nm range with two nanometre spacing, together with reference values for moisture, oil, protein, and starch[19]. The main purpose is to predict the protein content in corn in this study.

The integration of internal and public datasets ensured complementary coverage of application scenarios and spectral ranges. The internal datasets enabled targeted evaluation of classification and anomaly detection in food and traditional medicine, and supported a realistic regression task in wastewater analysis. Public benchmarks, by contrast, offered well documented provenance, stable sampling grids, and standardized reference targets, thereby enhancing reproducibility and enabling direct comparison with conventional machine learning and deep learning baselines under consistent preprocessing and evaluation protocols.

Table 1 Experimental material data summary

| Sample Name | Samples | Wavelength Range | Task Type | Note |
|---|---|---|---|---|
| Milk | 360 | 1100-2300nm | Classification, Anomaly Detection | 9 classes with different mixture ratios of regular vs lactose free milk |
| Chinese medicinal herbs | 120 | 900-1700nm | Classification, Anomaly Detection | 3 classes: Lonicerae Japonicae Flos (LJF), Lonicerae Flos (LF) and the mixture of them |
| Citri Reticulatae Pericarpium | 240 | 900-1700nm | Classification, Anomaly Detection | 8 classes with storage durations ranging from 1 to 8 years |
| Wastewater | 67 | 190-1100nm | Regression | Samples with different COD |
| Tecator | 215 | 850-1050 nm | Regression | Samples with different fat content |
| Corn | 80 | 1100-2498 nm | Regression | Samples with different protein content |

## 2.2 Structured Knowledge Base and Entity Retrieval

In order to endow our automated spectral analysis framework with domain specific knowledge, we built upon the Spectral Data Analysis and Application Platform (SDAAP) dataset developed in our prior work, which represents the first systematically curated, open access text corpus dedicated to spectroscopic analysis and detection[20]. SDAAP not only aggregates and annotates peer reviewed publications across diverse spectroscopic applications, but also supplies each record with structured "knowledge guidance" entries that specify experimental targets, spectral modalities, preprocessing protocols, feature extraction techniques, and modeling approaches. This resource addresses a critical deficiency in publicly available text datasets for spectroscopy and provides a solid foundation for subsequent LLM applications in this domain.

From the SDAAP repository, we filtered entries that were directly relevant to the materials investigated in this study, including milk dataset, traditional Chinese medicinal herbs, the Tecator and Corn datasets, and Citri Reticulatae Pericarpium, with the aim of aligning the literature knowledge base with the experimental datasets described in Section 2.1. Each selected publication was processed using MinerU, an open source automated literature mining tool[21], and the extracted annotations were subsequently integrated into a large language model pipeline to obtain structured metadata. The metadata encompassed the material or analyte under investigation, the spectroscopic technique and wavelength ranges employed, the optimal preprocessing workflow applied to the raw spectra, the most effective feature derivation strategy, and the predictive modeling approach adopted in each study.

Within our end-to-end spectral analysis system, user input comprises two modalities: a natural language query and raw spectral measurements. Upon receiving a query, the system invokes an LLM with a bespoke prompt engineered for entity recognition. The LLM parses the question to identify two key entities: the research object and the analysis task type (classification, anomaly detection, or regression). Once the research object entity is identified, retrieval algorithms are used to search the structured knowledge base for records whose material descriptors match or closely resemble the query entity. The retrieved structured entries are then forwarded, alongside the raw spectral data, to a downstream LLM agent module that synthesizes prior knowledge with the new measurements. Concurrently, the extracted task type entity informs a multi-task inference controller within the LLM agent, directing it to apply the appropriate reasoning paradigm whether discriminative classification, binary anomaly detection, or continuous value regression to produce the final analytical output. Figure 2 shows the specific process of entity extraction and knowledge retrieval.

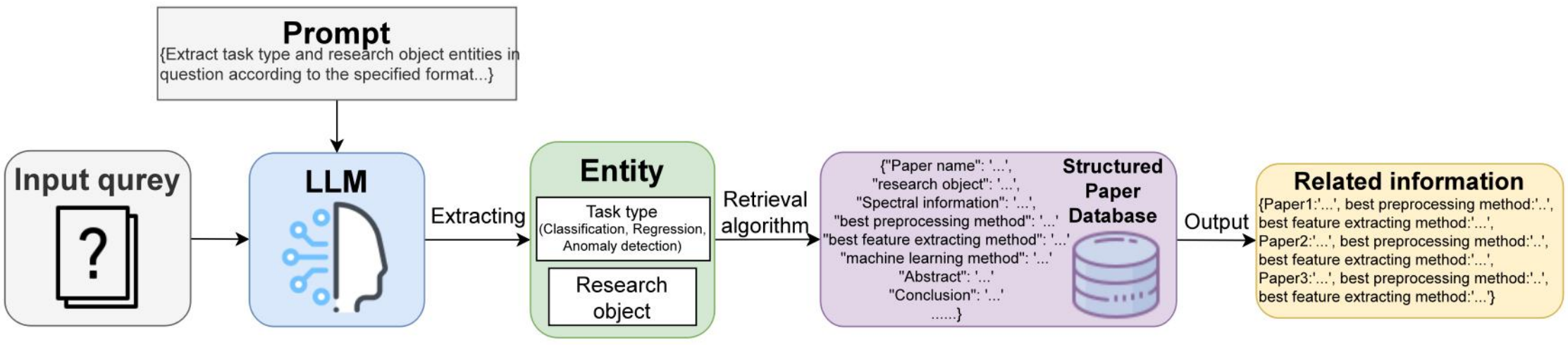

Figure 2. The specific process of entity extraction and knowledge retrieval

### 2.3 LLM Agent for Spectral Preprocessing and Feature Extraction

The LLM agent drives spectral data conditioning and representation by translating structured metadata from the knowledge base into a methodology selection framework. By default, upon receiving a raw absorbance spectrum and the extracted entity, the agent evaluates the top-k most relevant literature sources and determines the preprocessing and feature extraction pipeline through majority voting over these references. In cases where multiple pipelines appear with equal frequency, the system defaults to the first valid pipeline encountered in the ranked list. The selected algorithms from the spectral analysis library are then applied, outputs are validated against predefined quality assurance metrics, and, if necessary, the process is iterated until satisfactory spectral quality is achieved.

To provide transparency and user control, the interface displays the top-k relevant literature sources along with the methods reported in them. If the user is not satisfied with the automatically chosen pipeline, they can directly override the selection by specifying any one of the listed literature sources, in which case the agent immediately re-executes the processing with the corresponding method pair.

Baseline correction(BC) is offered via the asymmetric least squares (AsLS) algorithm, which estimates the corrected spectrum $z$ by minimizing[22].

$$\sum_{i=1}^{n}(y_i - z_i)^2 + \lambda \sum_{i=2}^{n}(z_{i+1} - 2z_i + z_{i-1})^2, \tag{1}$$

where $y_i$ denotes the raw intensity at index i, $z_i$the baseline subtracted intensity, and $\lambda$ a smoothing penalty that balances fidelity to the original spectrum against baseline smoothness.

Spectral smoothing is performed with the Savitzky-Golay(SG) filter, which fits a polynomial of degree over a moving window of width $2m + 1$ and computes the smoothed value[23].

$$\hat{y}_i = \sum_{j=-m}^{m} c_j y_{i+j} \tag{2}$$

where the convolution coefficients $c_j$ arise from the least squares fit of the local polynomial.

Normalization options include min-max scaling, which linearly rescales the intensities as

$$y_i^{'} = \frac{y_i - \min_k y_k}{\max_k y_k - \min_k y_k} \tag{3}$$

and the standard normal variate (SNV) transform[24], which centers and scales each spectrum by its own statistics:

$$y_i^{''} = \frac{y_i - \bar{y}}{\sigma_y} \tag{4}$$

with $\overline{y}$ and $\sigma_y$ representing the mean and standard deviation of the spectrum, respectively.

When derivative spectra are required, the first derivative is approximated by

$$y_i^{(1)} = \frac{y_{i+1} - y_{i-1}}{2\Delta\lambda} \tag{5}$$

and the second derivative by

$$y_i^{(2)} = \frac{y_{i+1} - 2y_i + y_{i-1}}{\Delta\lambda^2} \tag{6}$$

where $\Delta\lambda$ is the uniform wavelength increment.

After preprocessing, the agent proceeds to feature extraction. Options include principal component analysis (PCA), partial least squares (PLS) regression, continuous wavelet transform (CWT), Nonnegative Matrix Factorization (NMF) and a Lambert-Beer-Pearson correlation approach. In the fully automated workflow, the method most relevant to the input task type is executed; however, the user can override this choice via the same clickable method list shown in the interface. PCA is implemented by solving the eigenvalue problem.

$$Cu_k = \lambda_k u_k, \quad C = \frac{1}{n-1} X^\top X \tag{7}$$

where $X$ is the preprocessed data matrix, $u_k$ are the principal directions, and $\lambda_k$ their variances. PLS iteratively derives latent scores by maximizing covariance between the predictor matrix X and response vector y, extracting weight vectors w and loadings $p$, $q$ through

$$\max_{\|w\|=1} Cov(Xw, y), \quad X \leftarrow X - tp^\top, \quad y \leftarrow y - tc \tag{8}$$

where $t = \mathrm{Xw}$, $u = \mathrm{yq}$, and $c$ is the loading on y. The CWT is computed as

$$W(a,b) = \frac{1}{\sqrt{a}} \int_{-\infty}^{\infty} y(\lambda) \psi\left(\frac{\lambda - b}{a}\right) d\lambda \tag{9}$$

with scale $a$, translation $b$, and mother wavelet $\psi$.

In scenarios where a small set of reference spectra with known analyte concentrations is available,such as calibration of chemical oxygen demand or quantification of specific compound concentrations,the agent can instead invoke a hybrid Lambert-Beer-Pearson method[25]. Here, the absorbance at each wavelength $\lambda$ is first computed via the Lambert-Beer law,

$$A(\lambda) = \log \frac{I_0(\lambda)}{I(\lambda)} \tag{10}$$

where $I_0$ is the incident light intensity and I the transmitted intensity. These absorbance spectra are then compared against each reference curve $R_{ij}(\lambda)$ by computing the Pearson correlation coefficient

$$r_j = \frac{\sum_i (A_i - \bar{A})(R_{ij} - \bar{R}_j)}{\sqrt{\sum_i (A_i - \bar{A})^2} \sqrt{\sum_i (R_{ij} - \bar{R}_j)^2}} \tag{11}$$

The top n coefficients $\{r_j\}$ serve as features that capture both the linear absorbance behavior and the spectral similarity to calibration standards.

This design ensures that the pipeline is fully automatic by default while allowing rapid, transparent user overrides when domain expertise suggests a different approach. Figure 3 shows the specific process of preprocessing and feature extraction.

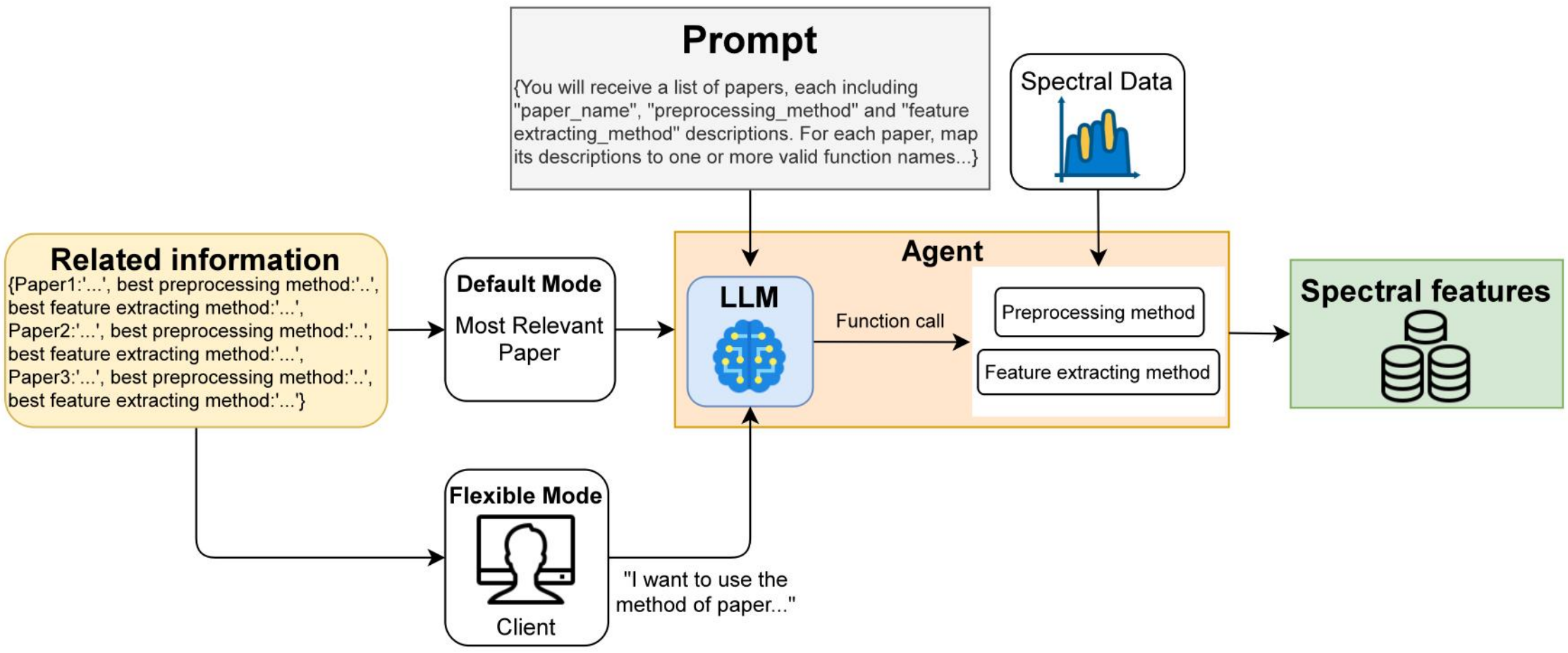

Figure 3. The specific process of preprocessing and feature extracting

## 2.4 LLM Multi-task Reasoning Pipeline

Classification, anomaly detection, and regression are performed by a large language model with fixed parameters, and behavior is governed by the prompt rather than by weight updates. For each dataset, a compact set of exemplar pairs from the training partition is assembled into an instruction oriented prompt that specifies task definition, spectral preprocessing choices, feature representations, and a small number of labeled exemplars. Given a query spectrum $x_i$, the model produces a categorical label for classification and anomaly detection or a scalar response for regression. No gradient based optimization is involved. All adaptation arises from information encoded in the prompt and from the curated exemplars that anchor task specific reasoning. Figure 4 illustrates the overall workflow.

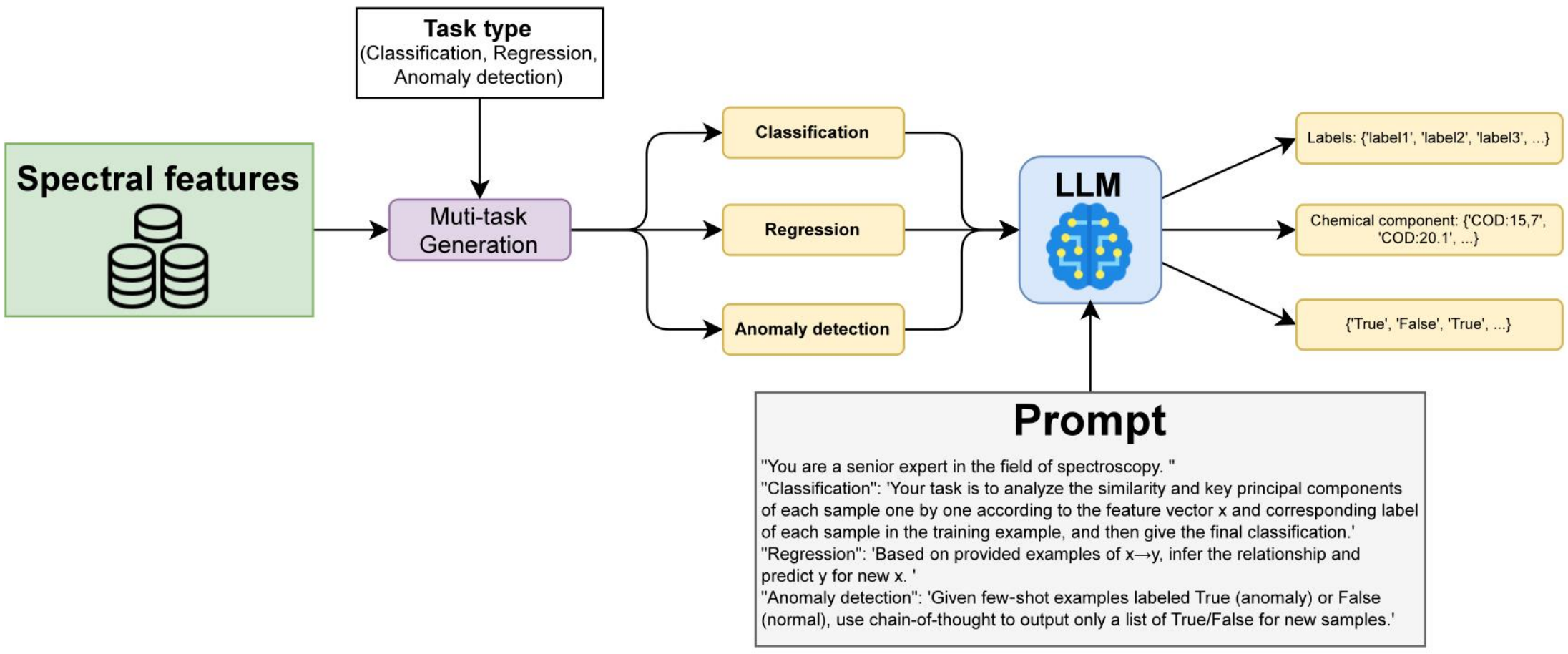

Figure 4. The specific process of LLM multi-task reasoning

Performance is evaluated on the held out test partition using task appropriate metrics. Let T denote the test set and $N_T = |T|$. For classification, accuracy is computed as

$$\mathrm{Acc} = \frac{1}{N_T} \sum_{i \in T} 1\,[\hat{y}_i = y_i] \tag{12}$$

For the anomaly detection task, the area under the receiver operating characteristic curve (AUC) is employed, defined as

$$\mathrm{AUC}^{(t)} = \frac{1}{N_{\mathrm{pos}} N_{\mathrm{neg}}} \sum_{i: y_i = 1} \sum_{j: y_j = 0} \mathbf{1}\left(s_i^{(t)} > s_j^{(t)}\right) \tag{13}$$

where $N_{pos}$ and $N_{neg}$ denote the numbers of anomalous and normal samples respectively, and $s_i^{(t)}$ the predicted anomaly score.

For regression, the coefficient of determination is reported as

$$R^2 = 1 - \frac{\sum_{i \in T} (\hat{y}_i - y_i)^2}{\sum_{i \in T} (y_i - \bar{y}_T)^2}, \quad \bar{y}_T = \frac{1}{N_T} \sum_{i \in T} y_i, \tag{14}$$

and root mean squared error is provided for completeness. The prompt templates used for inference can be serialized as JSON artifacts and reloaded verbatim to reproduce predictions under identical decoding settings, which underscores the distinction between prompt conditioned inference and parameter training.

This protocol differs fundamentally from machine learning and deep learning pipelines. In classical pipelines, data are consumed to estimate parameters through iterative optimization on a training split and validation is used for early stopping and hyperparameter selection. In the present setting, the foundation model retains its original weights and adapts through context alone. The few exemplars and task specifications embedded in the prompt act as conditioning evidence that steers domain specific reasoning at inference time. This leads to a distinct usage of data, since examples serve as contextual guidance rather than as gradient bearing samples. It also leads to a distinct cost profile, since compute is allocated to token processing rather than to parameter updates. Failure modes likewise differ. Parameterized models are sensitive to overfitting and underfitting, whereas prompt conditioned inference is sensitive to context length, exemplar selection, formatting, and decoding choices. For comparability, all approaches are evaluated on the same held out test partitions. Machine learning and deep learning baselines employ early stopping and hyperparameter search on the validation split, while the large language model performs inference without any parameter modification, and performance reflects only the information and constraints supplied by the prompt.

## 3. Results and discussion

### 3.1 Experiment settings

Experiments were conducted on a workstation equipped with an NVIDIA GeForce RTX 4070 Ti Super (16 GB). All large language model calls were issued via API. During entity extraction, qwen-turbo (2025-07-28) and qwen-plus (2025-07-15) were configured with temperature 0.1 to promote precise identification of research objects and task types.

For LLM based inference on spectra, temperature was fixed at 0.5 unless otherwise stated. Datasets followed a 6:2:2 split for training, validation, and testing. After initial benchmarking under this configuration, a backbone model was selected and applied across all datasets and tasks, with comparisons to standard machine learning and deep learning baselines. Ablation studies varied temperature and the number of training samples to quantify their effects on performance.

Preliminary trials indicated that regression labels often contain long decimal representations. When all samples are included, prompt length increases and can approach the context window limit, which may induce numeric hallucinations. To control context usage, the total sample count for regression experiments was capped at 100.

To ensure fair comparison, the 6:2:2 split was used consistently. The validation partition served as the early stopping criterion for deep learning baselines. Classical machine learning methods employed grid search for hyperparameter selection. The same training and test partitions were shared across LLM inference, machine learning baselines, and deep learning baselines so that performance differences reflect inference procedures rather than data selection. Each experiment was repeated ten times per material category, and results are reported as mean ± standard deviation.

## 3.2 Experiments of entity extraction and knowledge retrieval

### 3.2.1 Entity extraction accuracy

The performance of the entity extraction and subsequent knowledge retrieval modules was evaluated in order to assess the system's ability to identify both the research object and the analysis task from queries, and to retrieve the appropriate structured entries from the spectral literature corpus. For the entity extraction evaluation, a set of one hundred questions was manually constructed, covering three task types (classification, anomaly detection, regression) across five material categories. Each question was annotated with the ground truth research object label and task type label prior to testing. The prompts for entity extraction were then submitted to the qwen-turbo and qwen-plus models via API, with the research object entity evaluated using a fuzzy matching criterion (FuzzyWuzzy) and the task type entity assessed by exact match accuracy[26].

Table 2 Evaluation results of entity extraction

| Models | Acc(Research object) | Acc(Task type) |
|---|---|---|
| qwen-turbo | 82 | 89 |
| qwen-plus | 86 | 100 |

The results, summarized in Table 2, indicate that qwen-plus outperformed qwen-turbo on both metrics. Specifically, qwen-plus achieved an accuracy of 86% for extracting research object, compared to 82% for qwen-turbo, and correctly identified the task type in all cases (100% accuracy), whereas qwen-turbo attained 89% on this metric. These findings demonstrated that the enhanced language understanding capabilities of qwen-plus yield more precise entity recognition, particularly for distinguishing among closely related spectral materials.

Following successful entity extraction, the identified labels of research objects were used to query the structured knowledge base via retrieval algorithms. Manual inspection of a subset of retrieval outputs confirmed that over 90% of the top ranked documents contained metadata entries corresponding to the intended material, thereby validating the robustness of the retrieval pipeline. This end-to-end evaluation of entity extraction and knowledge retrieval establishes a reliable foundation for the downstream multi-task reasoning agents.

#### 3.2.2 **Retrieval effectiveness**

In this study, three established lexical retrieval models were employed to identify relevant publications from the structured corpus: the Bag of Words (BoW) model, BM25, and TF-IDF cosine similarity. The BoW model represents each document as an unordered multiset of word frequencies, disregarding syntactic structure and term dependencies[27]. While straightforward to implement, its inability to capture contextual nuances often yields suboptimal precision in specialized domains. BM25, or Best Matching 25, builds upon the probabilistic retrieval framework by incorporating term frequency saturation, document length normalization, and inverse document frequency weighting; it has become a de facto standard in information retrieval for its balanced treatment of term importance and document verbosity[28]. The TF-IDF cosine similarity approach combines classical term frequency inverse document frequency weighting with a vector space model, measuring the cosine of the angle between query and document vectors to quantify their alignment; this method effectively emphasizes domain specific vocabulary while mitigating the influence of overly common terms.

Retrieval performance was evaluated on a curated test corpus of 200 documents sampled from the full SDAAP repository. One hundred queries were derived from the standardized research object labels produced in Section 3.2.1. For each query, the three highest ranked documents were retrieved by each model and compared against a set of preassigned relevance labels. Retrieval precision was calculated as the fraction of correctly retrieved documents among the top three for each query, and the mean precision across all queries was reported.

Table 3 Evaluation results of knowledge retrieval

| Methods | Accuracy(%) |
|---|---|
| Bag of Words model | 56.00 |
| BM25 | 80.67 |
| TF-IDF cosine similarity retrieval method | 81.33 |

Table 3 summarizes the quantitative outcomes of these retrieval experiments. As shown, the Bag of Words model approach achieved a mean precision of 56.00%, reflecting the limitations of raw term counts in this specialized domain. The TF IDF model produced a substantially higher mean precision of 81.33%, demonstrating the benefit of inverse document frequency weighting in discriminating domain specific terms. BM25 attained a comparable mean precision of 80.67%, indicating that its term frequency saturation and document length normalization confer performance similar to TF IDF under these pilot conditions. Together, these results confirm that weighted retrieval schemes markedly outperform the unweighted Bag of Words approach in identifying relevant spectral analysis literature within a noisy corpus.

### 3.3 Spectral Preprocessing and Feature Extraction Outcomes

The preprocessing and feature extraction pipeline for each material follows literature guided routines summarized in Table 4. For the public Milk dataset, a SG smoothing step is applied before principal component analysis, while Chinese medicinal herbs use SNV followed by a FD filter prior to PCA, in order to enhance weak band structure and reduce scattering effects. Citri Reticulatae Pericarpium employs SNV with subsequent PCA for dimensionality reduction. The private wastewater dataset is baseline corrected, implemented as asymmetric least squares in our workflow, and then encoded as Pearson correlation features against reference chemical oxygen demand profiles. Tecator and Corn, both public datasets, are standardized with SNV and then represented by partial least squares latent variables for downstream modeling. These choices are exactly those recorded in the updated retrieval Table 4 for materials, preprocessing, and feature extraction methods.

Table 4 Retrieval results for experimental material related preprocessing methods and feature extraction methods

| Materials | Preprocess method | Feature extract method |
|---|---|---|
| Milk | SG | PCA |
| Chinese medicinal herbs | SNV+FD | PCA |
| CRP | SNV | PCA |
| Wastewater | BC | Pearson correlation features |
| Tecator | SNV | PLS |
| Corn | SNV | PLS |

Figure 5 illustrates the effect of these choices by juxtaposing raw and processed spectra with their corresponding feature maps. On Milk, smoothing suppresses high frequency noise and stabilizes the variance structure for PCA. In Chinese medicinal herbs, the combination of SNV and FD sharpening reveals subtle curvature associated with overlapping constituents. For Citri Reticulatae Pericarpium, SNV aligns baselines and improves comparability across samples, enabling principal components to capture the dominant chemical variance. In wastewater, removal of baseline drift by asymmetric least squares prepares the signal for correlation based compression relative to reference profiles. For Tecator and Corn, SNV precedes partial least squares, which yields a compact set of latent factors that summarize the predictive spectral subspace. Together these outcomes confirm that the metadata driven selection procedure produces well conditioned and low dimensional representations that are consistent with the updated table and optimized for subsequent inference.

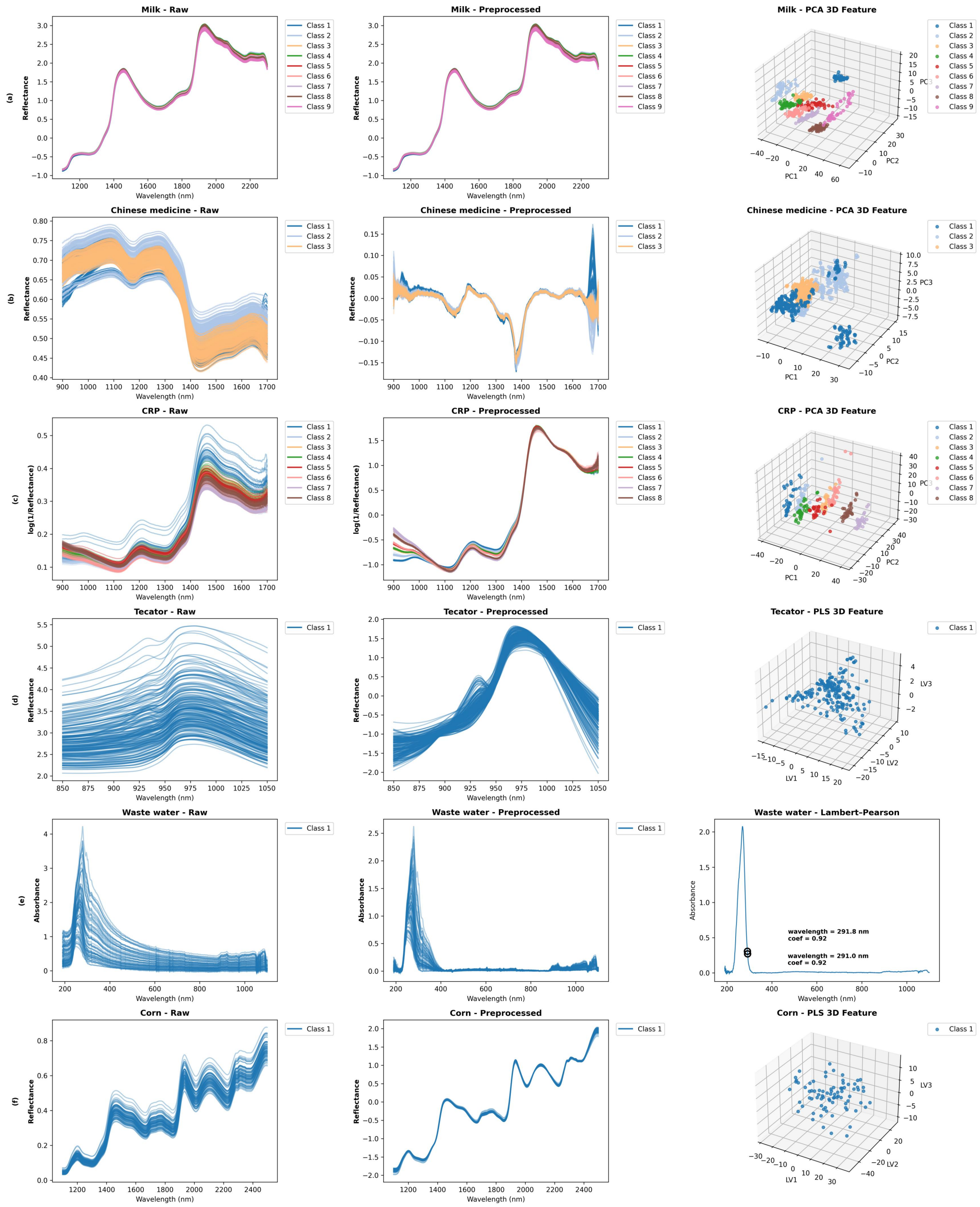

Figure 5. Raw spectra, preprocessed spectra, and resulting feature maps for six material categories: (a) Milk dataset, (b) Chinese medicinal herbs, (c) CRP (d) Tecator dataset, (e) Wastewater, (f) Corn dataset,

## 3.4 Experiments of multi-task reasoning

### 3.4.1 Model selection

This subsection identifies the default model for the proposed pipeline by jointly considering two complementary dimensions. Within this evaluation, four contemporary conversational models are examined. The Qwen series developed by Alibaba and accessed through the Alibaba Cloud API includes qwen-plus and qwen-turbo. Both adopt a Transformer backbone and are pretrained on large scale text and multimodal corpora to generate coherent, semantically rich outputs across domains. Qwen-plus, evaluated as the 2025-07-28 release, offers a balanced tradeoff among reasoning performance, cost, and latency, and is suitable for tasks of moderate complexity. Qwen-turbo, evaluated as the 2025-07-15 release, prioritizes inference speed and minimal cost, with correspondingly reduced reasoning depth for simpler workloads[29]. DeepSeek-v3, released on December 26, 2024 by DeepSeek, is a mixture of experts model with 671 billion parameters pretrained on 14.8 trillion tokens, targeting high capacity reasoning at scale[30]. ChatGPT-5, accessed as chatgpt-5-chat-latest from OpenAI, is a high capacity conversational model finetuned for instruction following and multi turn dialogue, and serves here as a strong general purpose baseline[31].

The first dimension comprises practical constraints that directly affect inference, including the effective context window exposed by each API, the cost per million tokens under provider pricing, and the average completion length required by each task. In this setting, classification yields short label strings, anomaly detection returns binary outputs, and regression produces substantially longer numeric sequences. To characterize token usage, the number of output tokens generated during inference was recorded. Results are presented in Table 5.

Table 5 Comparison of LLMs under generation constraints

| LLMs | Classification Output tokens | Regression Output tokens | Anomaly detection Output tokens | Context window limitations | Cost per million tokens |
|---|---|---|---|---|---|
| qwen-plus | 135 | 112 | 25 | 1000K | 0–128K：¥2，128–256K：¥20，256K–1M：¥48 |
| qwen-turbo | failed | 112 | 25 | 1000K | ¥6 |
| deepseek-v3 | 144 | 81 | 25 | 64K | ¥0.5 |
| chatgpt-5-chat | 146 | 81 | 26 | 400K | $10 |

The second dimension comprises empirical performance across tasks. For each model, outcomes were measured on shared dataset splits with matched prompts and temperature schedules, and the task metrics described above were summarized as means and dispersion across repeated runs. The benchmark includes qwen-plus, qwen-turbo, deepseek-v3, and chatgpt-5-chat-latest, evaluated on the same task suite, with Milk used for classification, Chinese medicinal herbs for anomaly detection, and Corn for regression. Experimental results are reported in Table 6. Instances marked as failed in Tables 5 and 6 are primarily attributable to the large sample size in the classification task, which exceeded the operational capacity of qwen-turbo in this setting and prevented the generation of valid classification outputs.

Table 6 Experiment results of different LLMs

| LLMs | Classification (Milk) | Regression (Corn) | | Anomaly detection (Chinese medicinal herbs) | |
|---|---|---|---|---|---|
| | Acc (%) | R2 | RMSE | Precision | AUC |
| qwen-plus | 69.72±3.7 | 0.8117±0.0351 | 0.2258±0.0214 | 1±0 | 1±0 |
| qwen-turbo | failed | 0.7320±0.0529 | 0.2693±0.0261 | 1±0 | 0.75±0 |
| deepseek-v3 | 50.83±5.69 | 0.6731±0.0588 | 0.2976±0.0274 | 1±0 | 1±0 |
| **chatgpt-5-chat** | **96.10±1.52** | **0.9252±0.0065** | **0.1427±0.0063** | **1±0** | **1±0** |

Across all three tasks ChatGPT-5 provides the most robust and accurate performance. Although its unit price per generated token is higher, the predictions studied here require very few tokens because outputs are short labels, compact numbers, or binary decisions. The effective cost per example is therefore modest while accuracy remains consistently high. Consequently, ChatGPT-5 is selected as the default model in the pipeline, and all subsequent experiments and ablations use ChatGPT-5 for inference.

### 3.4.2 Results of Classification

This task aims to evaluate the capability of different models to accurately classify spectroscopic samples from three material categories: milk, Chinese medicinal materials, and Citri Reticulatae Pericarpium (CRP). Each sample's input comprises raw spectral data subjected to category specific optimal preprocessing techniques, followed by feature extraction using PCA with five retained components. These low dimensional spectral features were used as inputs for both LLM based models and classical machine learning classifiers.

As baselines, three classical classifiers were tested: support vector machine (SVM)[32], k-nearest neighbors (KNN)[33], and random forest (RF) which are all trained on the same features extracted by PCA[34]. We further evaluated a one dimensional convolutional neural network (CNN)[35] and a simple Transformer encoder[36], both trained on the raw, preprocessed spectral vectors.

Table 7 Experiment results of classification task

| Methods | Milk Acc.(%) | Chinese medicinal herbs Acc.(%) | CRP Acc.(%) |
|---|---|---|---|
| LLM | 96.10±1.52 | 100±0 | 99.58±0.93 |
| SVM | 100±0 | 100±0 | 95.83±0 |
| KNN | 93.06±0 | 100±0 | 89.58±0 |
| RF | 97.22±0 | 100±0 | 95.83±0 |
| CNN | 95.83±0.98 | 90±14 | 91.25±1.75 |
| Transformer | 95.11±3.72 | 85±15 | 97.50±3.42 |

As shown in Table 7, across the three datasets, distinct performance patterns emerge among classical machine learning models, deep architectures, and the large language model (LLM). On the Milk dataset, SVM achieves perfect accuracy, while the LLM and random forest also perform competitively. CNN and Transformer yield slightly lower but still strong results, whereas KNN lags behind, reflecting its sensitivity to data distribution complexity.

For Chinese medicinal herbs, the task appears relatively simple, as most methods (LLM, SVM, KNN, RF) achieve perfect classification. By contrast, CNN and Transformer exhibit noticeable variability, suggesting that high capacity neural architectures are prone to instability under limited data conditions.

On the CRP dataset, the LLM attains the highest accuracy, surpassing both classical and deep learning baselines. Transformer also performs strongly, whereas SVM and RF deliver moderate results, and KNN again underperforms .

Taken together, these results indicate that the LLM consistently provides robust and accurate predictions across tasks, excelling particularly in more complex scenarios such as CRP. While SVM and RF remain competitive in simpler settings, their advantages diminish when nonlinearities increase. CNN and Transformer demonstrate potential but exhibit higher variance, reflecting a trade off between capacity and stability in small to medium scale data regimes.

#### 3.4.3 **Results of Regression**

In the regression study we evaluate linear regression, support vector regression[37], and partial least squares regression on reduced features[38], together with random forest and two end to end neural baselines, a one dimensional convolutional network and a compact Transformer, trained on preprocessed spectra. The large language model conducts direct single shot inference from exemplar augmented prompts without any parameter updates. For Tecator and Corn, which are public datasets, features are obtained through partial least squares with four components, while the wastewater dataset, which is private, uses a small set of physically interpretable features derived from absorbance laws and correlation structure. Table 8 summarizes the results in terms of coefficient of determination and root mean squared error.

Table 8 Experiment results of regression task

| Type | Methods | R2 | RMSE |
|---|---|---|---|
| Tecator | LLM | 0.9592±0.0033 | 2.9221±0.1180 |
| | LR | 0.9696±0 | 2.5234±0 |
| | SVR | 0.9876±0 | 1.6100±0 |
| | PLSR | 0.9696±0 | 2.5234±0 |
| | RF | 0.9551±0 | 3.0674±0 |
| | CNN | 0.9796±0.0062 | 2.0486±0.3117 |
| | Transformer | 0.8909±0.0726 | 4.5784±1.5438 |
| Corn | LLM | 0.9252±0.0065 | 0.1427±0.0063 |
| | LR | 0.8826±0 | 0.1789±0 |
| | SVR | 0.9096±0 | 0.1570±0 |
| | PLSR | 0.8850±0 | 0.1771±0 |
| | RF | 0.6474±0 | 0.3101±0 |
| | CNN | 0.7938±0.0365 | 0.2364±0.0212 |
| | Transformer | 0.7262±0.0874 | 0.2708±0.0411 |
| Wastewater | LLM | 0.9185±0.0066 | 12.2362±0.5011 |
| | LR | 0.7409±0 | 21.8310±0 |
| | SVR | 0.8277±0 | 17.8029±0 |
| | PLSR | 0.7447±0 | 21.6732±0 |
| | RF | 0.8013±0 | 19.1181±0 |
| | CNN | 0.7535±0.0804 | 21.0832±3.3542 |
| | Transformer | 0.9021±0.0655 | 12.8519±4.3135 |

Across datasets the ranking of methods is consistent with the interplay between feature design, sample size, and nonlinearity. On Tecator the strongest performance is achieved by a kernel based method, indicating that the residual mapping after partial least squares reduction remains smooth yet meaningfully nonlinear in a way that a radial basis function can capture with low variance. The convolutional baseline is also competitive, which suggests that local spectral patterns are stable enough for localized filters to generalize well. Linear models on the reduced space remain adequate but not optimal, and the Transformer shows the widest variability, a sign of sensitivity to initialization and limited data when model capacity is high. The large language model is close to but not at the top on this dataset, which is consistent with a setting where kernel smoothing is well matched to the structure of the remaining nonlinearity.

On Corn the large language model attains the best overall balance of higher coefficient of determination and lower error, outperforming both linear baselines and the kernel method on the same reduced features. This pattern implies that the mapping in the reduced space departs from linearity in a way that is better captured by the prompt conditioned function learned from exemplars. End to end deep models underperform in this regime, which points to a data requirement that is not fully met when training directly on full spectra. The stability of the large language model across repeats further indicates that exemplar conditioning on low dimensional, domain aware features can deliver both accuracy and robustness without parameter tuning.

On the private wastewater dataset the large language model again leads, followed by the Transformer, while classical baselines are clearly behind. Here the features encode physical constraints and pairwise relations, and the advantage of the large language model suggests that its prompt conditioned mapping can approximate the moderate nonlinearity expressed by these interpretable features more effectively than linear predictors and with less variance than a high capacity Transformer. Random forest and the convolutional baseline do not benefit as much from these compact features, which is consistent with their preference for richer representation learning directly from high dimensional inputs.

Taken together these observations offer a simple practical recipe. When domain informed reduction yields compact and informative features, direct large language model inference is a strong default that combines accuracy with stability and very low engineering overhead. If validation indicates a very smooth residual structure in the reduced space, a kernel method can still be preferable. End to end neural models remain valuable when data are abundant and the goal is to exploit fine grained structure in raw spectra, but in the small to medium sample regime they are more sensitive to initialization and regularization. The contrast between the two public datasets and the private wastewater dataset also highlights the role of feature construction. Public benchmarks with partial least squares features can favor methods that capture smooth nonlinearities with small variance, whereas physically motivated features extracted from domain principles can amplify the gains of exemplar conditioned large language model inference.

#### 3.4.4 Results of Anomaly Detection

This task investigates anomaly detection across three spectroscopic material categories: milk, Chinese medicinal herbs, and Citri Reticulatae Pericarpium (CRP). The input for each sample comprises spectroscopic data that has been preprocessed using the optimal spectral transformation for its material class, followed by dimensionality reduction via PCA with five components. This results in a five dimensional spectral feature vector per sample, which serves as the input to all models under evaluation.

For each material, a subclass was designated as the reference class and considered "normal," comprising 50% of the data. The remaining 50% was divided into two types of anomalies: 33.33% were taken from different subclasses (inter class anomalies), and 16.67% were generated by applying small perturbations to the reference data (intra class anomalies). All normal samples were labeled as True, while both anomaly types were labeled False. The model's task is to perform binary classification for each spectrum, indicating whether it conforms to the learned "normal" distribution. Evaluation metrics included precision and the area under the receiver operating characteristic curve (AUC), which together reflect detection accuracy and robustness under class imbalance.

As baselines, three classical one class anomaly detection models were also considered: Isolation Forest (IF), which uses random partitioning to isolate anomalies[39] and one class support vector machine (OC-SVM), which attempts to separate the origin from the data in feature space using a hyperplane[40]. To evaluate lightweight deep learning alternatives, we also tested an autoencoder trained on the normal class and an LSTM autoencoder cascade[41][42], both using the same spectral features.

Table 9 Experiment results of anomaly detection task

| Methods | Milk | | Chinese medicinal herbs | | CRP | |
|---|---|---|---|---|---|---|
| | Precision | AUC | Precision | AUC | Precision | AUC |
| LLM | 1±0 | 0.950±0.046 | 1±0 | 1±0 | 1±0 | 0.75±0 |
| IF | 0±0 | 0.500±0 | 0±0 | 0.500±0 | 0±0 | 0.5±0 |
| OC-SVM | 1±0 | 0.167±0 | 0.625±0 | 0.2778±0 | 1±0 | 0.125±0 |
| Autoencoder | 1±0 | 1±0 | 1±0 | 0.628±0.061 | 1±0 | 0.975±0.056 |
| LSTM+Autoencoder | 1±0 | 0.994±0.012 | 1±0 | 0.628±0.046 | 1±0 | 0.75±0 |

The results, summarized in Table 9, reveal distinct performance patterns across datasets and methods. The LLM demonstrates consistently high precision across all three materials, with particularly strong performance on Chinese medicinal herbs where both precision and AUC reach perfect scores even when the number of training samples is limited. This robustness suggests that the separation between normal and anomalous samples in this dataset is well captured by the LLM representations, allowing reliable detection even under minimal supervision. For the milk dataset, the LLM also achieves perfect precision and a stable AUC close to 0.95, outperforming conventional approaches that fail to detect anomalies reliably. On the CRP dataset, while precision remains perfect, the AUC is lower, indicating that although the model can classify anomalies correctly, its ability to rank samples by anomaly likelihood is more limited in this case.

Among the baselines, IF and OC-SVM exhibit poor detection capability, often defaulting to random or biased predictions. In contrast, the autoencoder achieves strong results on milk and CRP, showing that reconstruction-based strategies can capture meaningful deviations in spectral structure. However, its performance drops markedly on Chinese medicinal herbs, highlighting the dataset-dependent sensitivity of such approaches. The LSTM–autoencoder cascade improves the stability of AUC on milk but shows no substantial advantage on the other datasets, suggesting that increased model complexity does not necessarily translate into more effective anomaly detection in this domain.

Overall, these findings confirm the ability of the LLM framework to deliver robust and accurate anomaly detection across diverse spectroscopic materials. Its superiority over traditional machine learning methods and competitive performance relative to specialized deep architectures underscore the value of leveraging language-model-based reasoning for complex chemometric tasks.

## 3.5 Ablation study

This section examines two design factors of the proposed agent—decoding temperature and the size of the training set—across three representative tasks: Milk classification, Corn regression, and anomaly detection on Chinese medicinal herbs. All trends are discussed with respect to Table 10 and Table 11.

### 3.5.1 **Temperature**

Temperature modulates the entropy of the LLM's token sampling and therefore governs the balance between determinism and exploratory diversity during reasoning. Table 10 reports the experimental results obtained under different temperature settings, covering classification, regression, and anomaly detection tasks.

Table 10 Experiments results of different temperature

| Temperature | Classification (Milk) | Regression (Corn) | | Anomaly detection (Chinese medicinal herbs) | |
|---|---|---|---|---|---|
| | Acc (%) | R2 | RMSE | Precision | AUC |
| 0.1 | 86.10±6.21 | 0.9354±0.0044 | 0.1326±0.0046 | 1±0 | 1±0 |
| 0.3 | 96.66±0.76 | 0.9344±0.0081 | 0.1336±0.0083 | 1±0 | 1±0 |
| 0.5 | 96.10±1.52 | 0.9252±0.0065 | 0.1427±0.0063 | 1±0 | 1±0 |
| 0.7 | 89.16±18.01 | 0.9300±0.013 | 0.1377±0.0131 | 1±0 | 1±0 |
| 0.9 | 87.77±16.50 | 0.9192±0.0135 | 0.1480±0.0127 | 1±0 | 1±0 |

Table 10 illustrates how temperature affects model performance across classification, regression, and anomaly detection tasks. For Milk classification, accuracy improves as temperature increases from 0.1 to 0.3, reaching its best performance around 0.3-0.5, but then declines at higher temperatures. This pattern indicates that moderate randomness in sampling helps balance stability and exploration, while excessive determinism or stochasticity undermines classification robustness.

For Corn regression, the best results are observed at low temperatures. At 0.1, the model achieves the highest R2 and lowest RMSE, while performance degrades gradually as temperature increases, with R2 dropping and RMSE rising at 0.9. This trend highlights that regression tasks favor more deterministic outputs, as randomness introduces variance that weakens continuous value prediction.

In Chinese medicinal herbs anomaly detection, Precision and AUC remain consistently perfect across all temperatures, suggesting that the task is insensitive to sampling variations, likely due to clear separability in feature space.

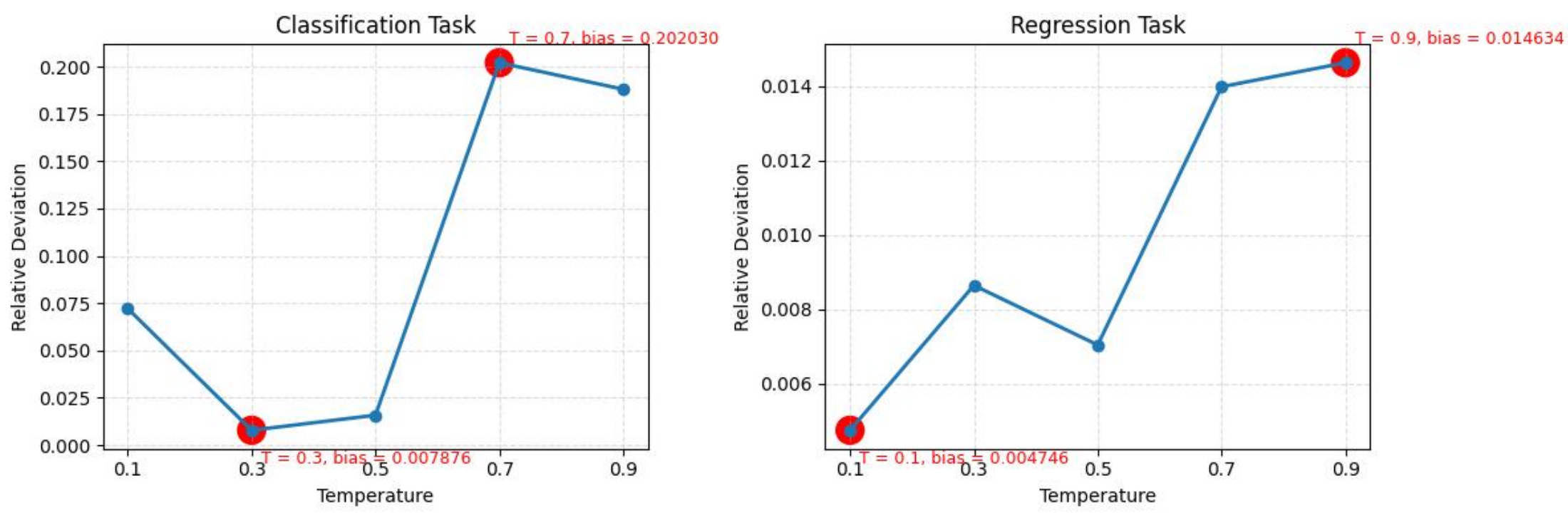

Figure 6. Temperature ablation analysis of relative average deviation (RAD) for classification and regression tasks.

To further validate the impact of temperature, we additionally plot the relative average deviation (RAD) curves in Figure 6. For classification, the RAD reaches its minimum at T = 0.3 and maximum at T = 0.7, showing that moderate temperatures stabilize the model's predictions while higher temperatures lead to unstable deviations. For regression, RAD is lowest at T = 0.1 and highest at T = 0.9, reinforcing that regression tasks require lower temperatures to minimize prediction variance. Together, the RAD analysis quantitatively confirms that classification benefits from moderate stochasticity, regression favors low temperatures, and anomaly detection remains stable regardless of temperature.

Taken together, these findings suggest that optimal temperature settings depend on task type: classification benefits from moderate value, regression favors lower values, and anomaly detection remains robust regardless of temperature.

### 3.5.2 Size of dataset

Varying the number of training samples reveals distinct data efficiency profiles for the three tasks, while keeping the test set size fixed during experiments. Table 11 presents the impact of varying training set sizes on model performance for classification, regression, and anomaly detection.

Table 11 Experiment results of different size of dataset

| Task | Size of Training set | Metrics | |
|---|---|---|---|
| Classification (Milk) | | Acc (%) | |
| | 72 | 30.83±3.85 | |
| | 144 | 47.50±3.00 | |
| | 216 | 96.10±1.52 | |
| Regression (Corn) | | R2 | RMSE |
| | 16 | 0.8354±0.0236 | 0.2114±0.0152 |
| | 32 | 0.8472±0.0281 | 0.2035±0.0184 |
| | 48 | 0.9252±0.0065 | 0.1427±0.0063 |
| Anomaly detection (Chinese medicinal herbs) | | Precision | AUC |
| | 6 | 0.75±0 | 0.8333±0 |
| | 12 | 1±0 | 1±0 |
| | 24 | 1±0 | 1±0 |
| | 36 | 1±0 | 1±0 |

Table 11 presents the impact of training set size on classification, regression, and anomaly detection tasks. For Milk classification, accuracy increases sharply with larger datasets. This steep gain highlights the critical role of sufficient data for learning reliable class boundaries in spectral classification.

For Corn regression, performance also improves with additional training samples. These results suggest that regression benefits strongly from larger datasets, as more samples enable better approximation of continuous relationships.

In contrast, Chinese medicinal herbs anomaly detection is largely insensitive to labeled data size: even with only six training samples (3 normal samples, 2 inter class anomalies and 1 intra class anomalies), it already delivers strong performance with high Precision and AUC, and the metrics quickly approach saturation as more data are added. This robustness likely reflects pronounced separability between normal and anomalous spectra in the learned feature space, enabling a reliable decision rule from only a handful of examples.

Overall, these results reveal task dependent sensitivity to training set size: classification and regression performance depend heavily on dataset scale, while anomaly detection appears inherently robust and stable regardless of training size.

## 4. Conclusion

We have presented an LLM-Driven Unified Agent framework for multi-task Reasoning in Infrared Spectroscopy that seamlessly combines literature guided preprocessing, feature extraction, and multi-task reasoning. By harnessing few-shot prompting, the system attains robust performance on classification, regression, and anomaly detection across heterogeneous spectral datasets, outperforming or matching traditional machine learning and simple deep learning baselines with minimal training data. Ablation studies underscore the roles of LLM sampling temperature, dataset size, and model selection in optimizing inference quality.

Future work will extend this agent architecture to accommodate image based spectral modalities, incorporating modules for multispectral and hyperspectral imaging. By integrating spatial-spectral feature extraction and domain specific LLM prompting, the enhanced framework will address complex analytical scenarios, such as material mapping and remote sensing, further broadening the applicability of intelligent, automated spectral analysis.

## CRediT authorship contribution statement

**Zujie Xie:** Methodology, Data curation, Validation, Formal analysis, Visualization, Software, Writing. **Zixuan Chen:** Data curation, Validation, Formal analysis, Visualization, Writing. **Jiheng Liang:** Conceptualization, Methodology. **Xiangyang Yu:** Conceptualization, Supervision, Writing. **Ziru Yu:** Conceptualization, Methodology, Supervision, Writing.

## Funding Sources

This research did not receive any specific grant from funding agencies in the public, commercial, or not-for-profit sectors.